\documentclass[10pt,twocolumn,letterpaper]{article}

\usepackage{cvpr}              

\usepackage{enumitem}
\usepackage{xspace}
\usepackage{ifthen}
\usepackage[normalem]{ulem}

\newcommand{\afterfigure}{\vspace*{-0.4cm}}

\newcommand{\SB}[2][]{%
    \ifthenelse{ \equal{#1}{} }
        {\textcolor{orange}{#2\xspace{}}}
        {\textcolor{orange}{\sout{#1} #2\xspace{}}}
}

\newcommand{\Ido}[2][]{%
    \ifthenelse{ \equal{#1}{} }
        {\textcolor{blue}{#2\xspace{}}}
        {\textcolor{blue}{\sout{#1} #2\xspace{}}}
}

\newcommand{\figTeaser}{
\twocolumn[{
\renewcommand\twocolumn[1][]{##1}
\maketitle
\centering
\vspace*{-8mm}
\includegraphics[width=.87\linewidth]{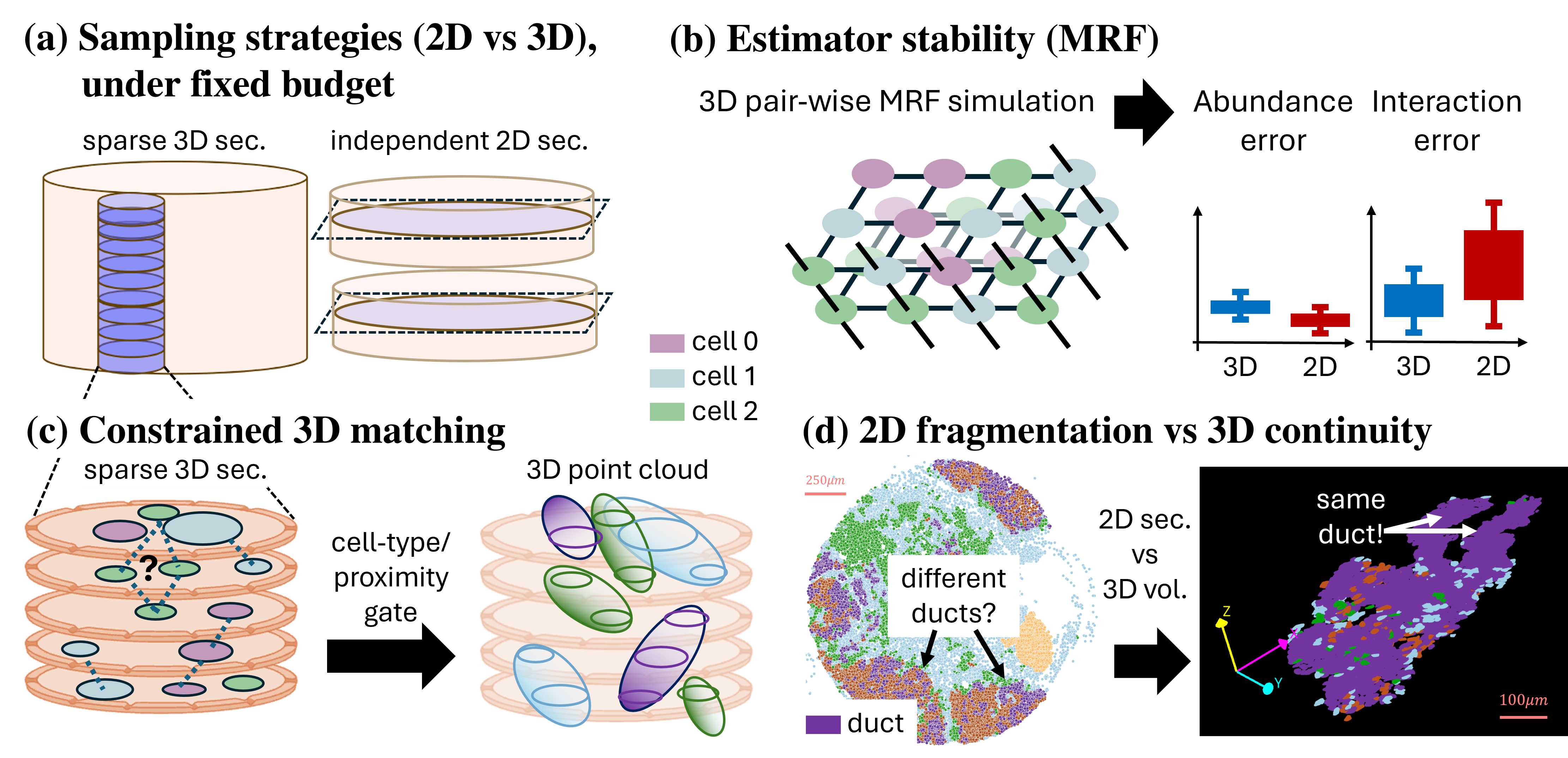}
\captionof{figure}{
\textbf{Sampling geometry governs estimator stability and geometric measurements under fixed imaging budgets.}
(\textbf{a}) Under the same imaging area, practitioners choose between sparse serial sections (depth continuity) and independent 2D sections (coverage).
(\textbf{b}) Using a pairwise MRF simulation framework, we show that global prevalence (abundance) is stably recovered under both strategies, whereas local interaction structure exhibits substantially higher error under planar sampling.
(\textbf{c}) We introduce a geometry-aware reconstruction module that formulates inter-slice correspondence as constrained assignment with phenotype and proximity gating.
(\textbf{d})~2D projection induces structural fragmentation,e.g., of the purple ducts, whereas 3D reconstruction restores connectivity.
}
\label{fig:teaser}
\vspace*{3mm}
}]
}

\newcommand{\figSim}{
\begin{figure}
\centering
\includegraphics[width=\linewidth]{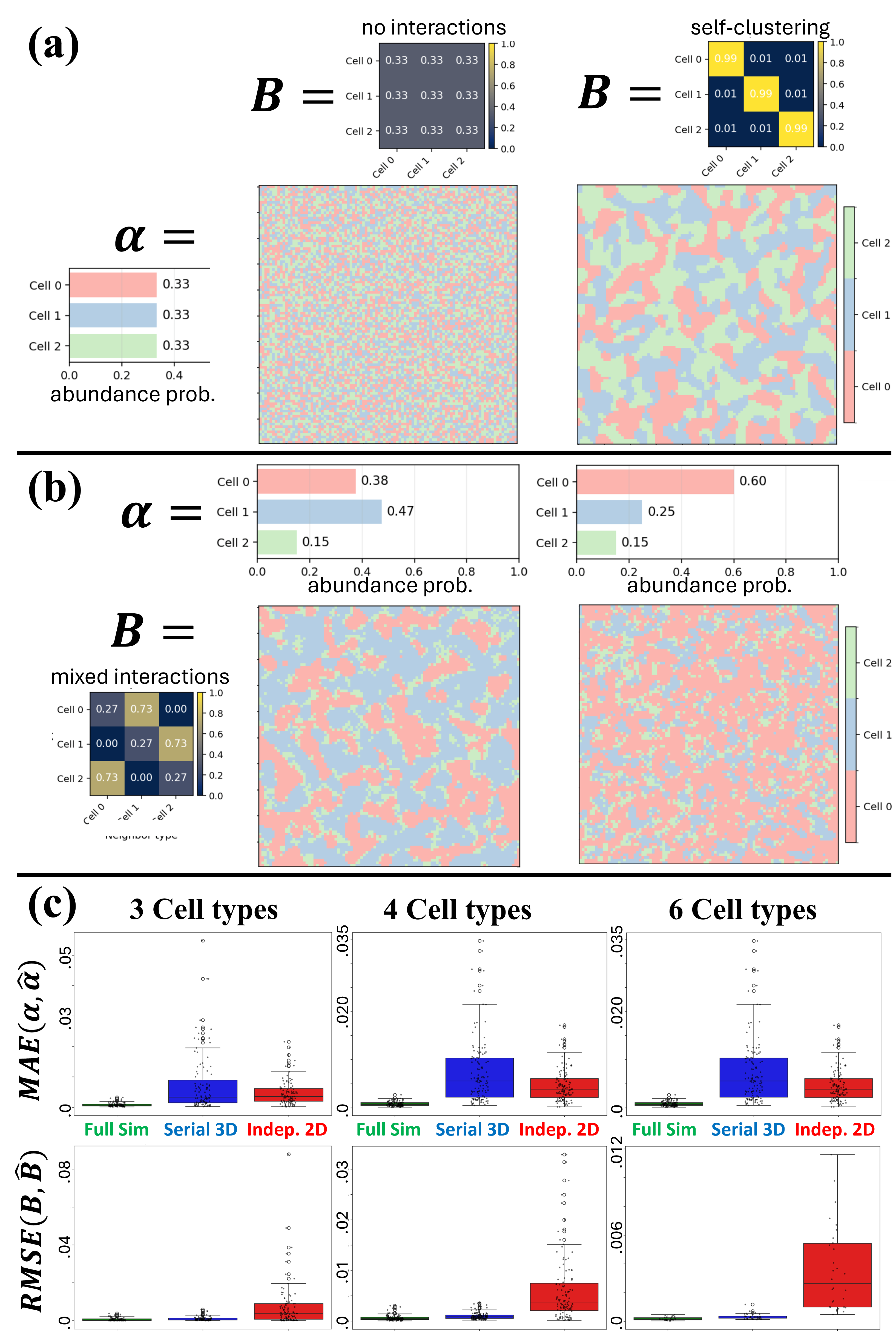}
\caption{
\textbf{Simulation (MRF): sampling geometry affects recovery of global prevalence versus local interactions.}
We simulate 3D tissues from a pairwise MRF with unary parameters $\boldsymbol{\alpha}$ (global prevalence) and pairwise affinities $\mathbf{B}$ (local interactions).
(\textbf{a})~With the same $\boldsymbol{\alpha}$, changing $\mathbf{B}$ yields markedly different spatial organizations, from near-i.i.d.\ mixtures to clustered domains. 
(\textbf{b})~Changing the $\boldsymbol{\alpha}$, for a fixed $\mathbf{B}$ yields different scales of structures.
(\textbf{c})~Under matched voxel budgets, MPLE recovers $\boldsymbol{\alpha}$ reliably from both independent 2D sections and sparse serial sampling, whereas recovery of $\mathbf{B}$ (and interaction-derived quantities) exhibits substantially higher error and variance under independent 2D sampling, especially in regimes with spatial localization.
\afterfigure}
\label{fig:sim}
\end{figure}
}

\newcommand{\figReal}{
\begin{figure}
\centering
\includegraphics[width=.93\linewidth]{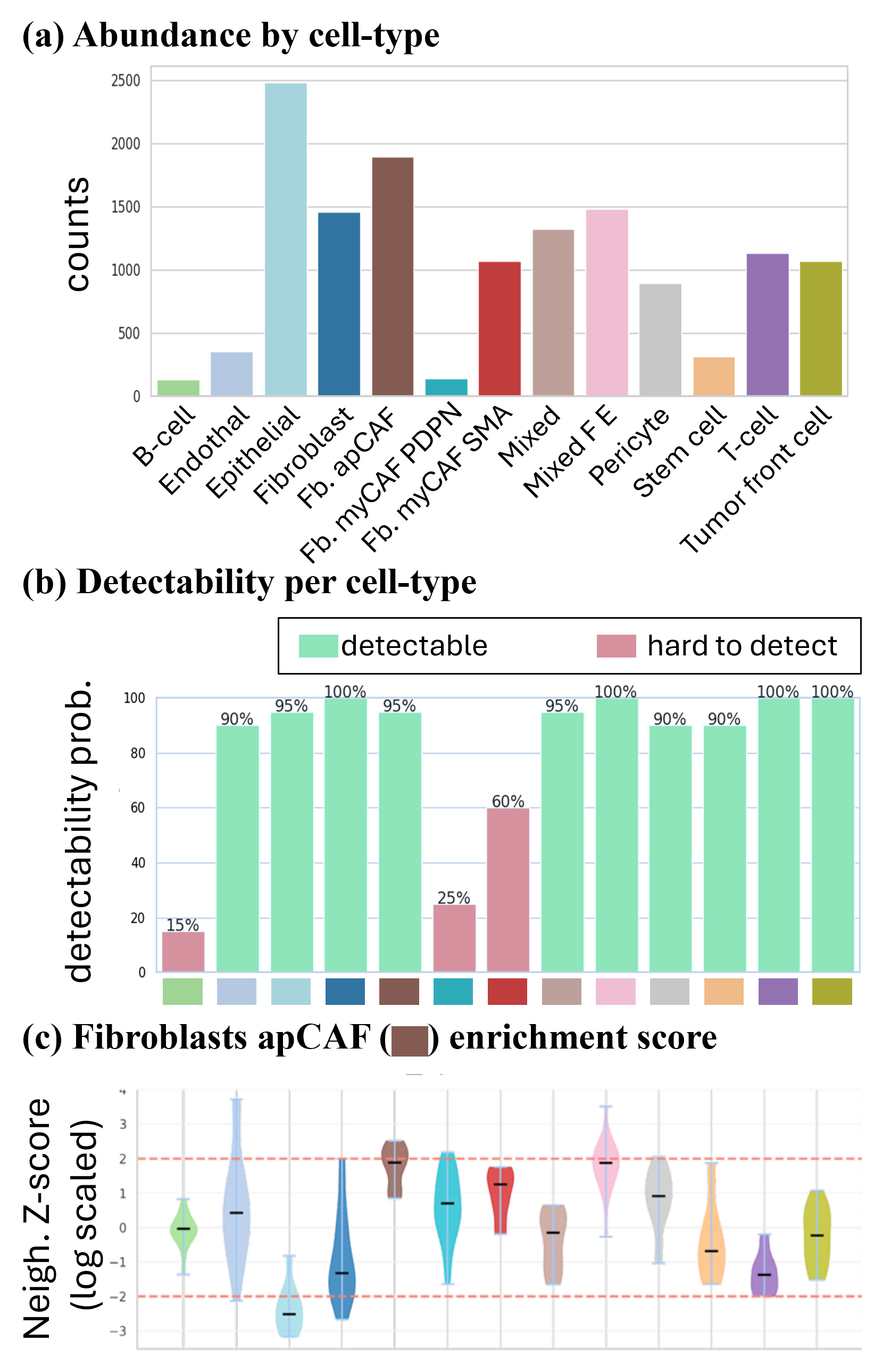}
\vspace*{-0.4cm}
\caption{
\textbf{Empirical stability in real IMC under dense axial sampling (2~$\mu$m;~\cite{kuett2022imc}).}
We treat a densely sampled IMC stack as a volumetric reference and repeatedly evaluate section-based statistics across depth.
(\textbf{a}) Total cell counts per type in the full volume (global abundance).
(\textbf{b}) \emph{Detectability} under planar sampling: for each type, the probability that a random set of $M{=}20$ sections contains at least $k{=}100$ cells of that type (a section-level proxy for 3D dispersion vs clustering).
B cells have low detectability, due to their low abundance and tendency to cluster, while similarly low-abundance but spatially dispersed types (e.g., endothelial) are detected consistently; conversely, abundant yet clustered types (e.g., myCAF\_SMA fibroblasts) can still be missed in many section samples.
(\textbf{c}) Variability of neighborhood enrichment for Fibroblasts\_apCAF: distribution (across sampled sections) of enrichment z-scores against each partner type (dashed lines indicate $|z|{=}2$).
Even within a single tissue volume, section choice can change whether an interaction appears strongly enriched or negligible, illustrating high sampling-induced variance. 
\afterfigure}
\label{fig:imc}
\end{figure}
}

\newcommand{\figMatching}{
\begin{figure}
\centering
\includegraphics[width=1.\linewidth]{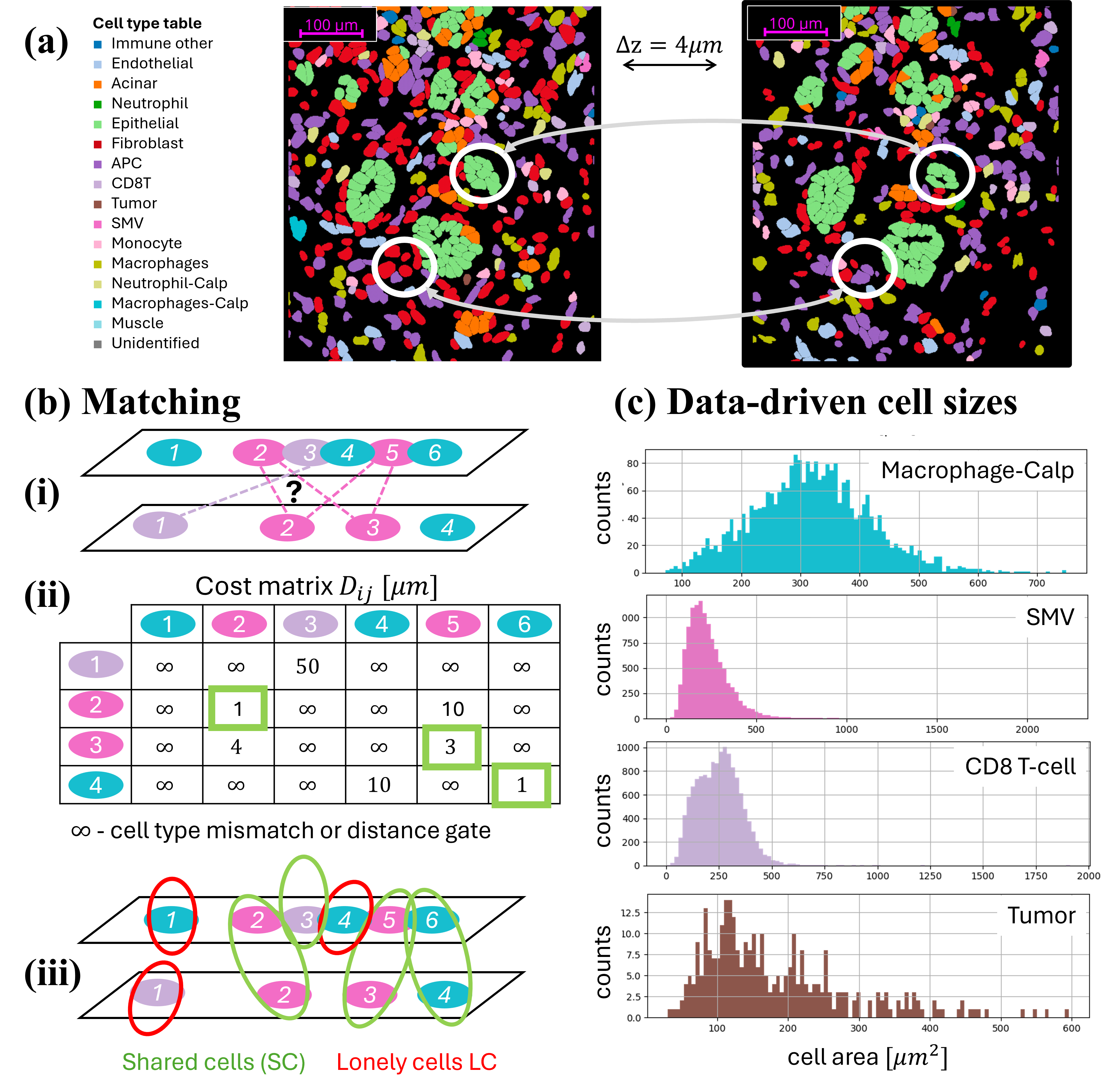}
\caption{
\textbf{Inter-slice matching as constrained assignment with cell-type-specific geometric priors (PDAC CODEX, $\Delta z=4\,\mu$m).}
(\textbf{a}) Two consecutive sections illustrate correspondence ambiguity: multiple nearby candidates of the same phenotype may exist for a given cell projection (highlighted regions).
(\textbf{b})(i)~Candidate correspondences are constructed under two constraints: phenotype consistency and a cell-type-specific proximity gate.
(ii)~We form the in-plane distance matrix $D_{ij}$ [$\mu m$] between cells in adjacent sections; entries violating phenotype or distance constraints are set to $\infty$.
Green dashed entries indicate selected matches after solving a one-to-one assignment via the Hungarian algorithm.
(iii)~The assignment partitions observations into \emph{shared cells} (SC), matched across adjacent sections; and \emph{lone cells} (LC), observed only in one section.
(\textbf{c}) Cell-type-specific proximity tolerances and weak ellipsoidal priors are derived from empirical distributions of 2D cross-sectional areas (examples shown).
\afterfigure}
\label{fig:method}
\end{figure}
}

\newcommand{\figMatchingRes}{
\begin{figure}
\centering
\includegraphics[width=\linewidth]{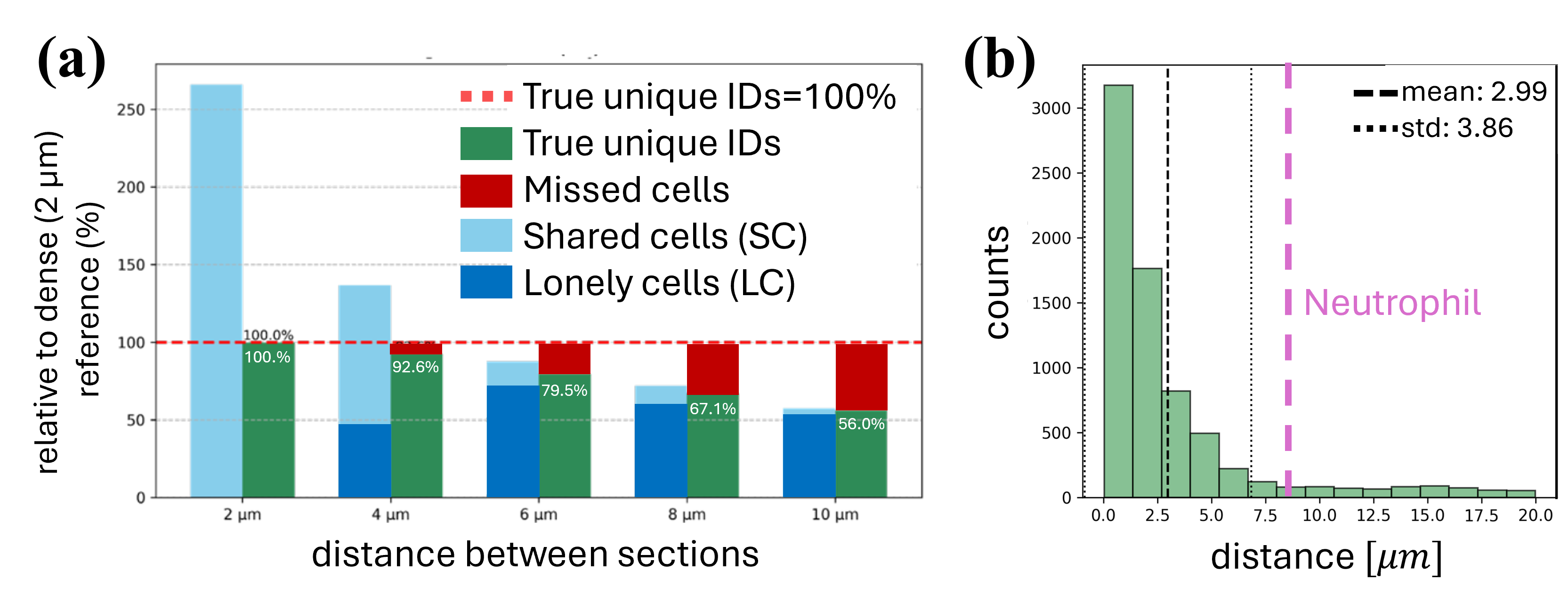}
\caption{
\textbf{Spacing trade-offs and localization accuracy on densely sampled IMC data~\cite{kuett2022imc}.}
(\textbf{a}) Effect of axial spacing $\Delta z$ relative to a dense 2~$\mu$m reference.
For each spacing, the left stacked bar shows the composition of cells \emph{observed in the sampled stack}, decomposed into shared cells (SC; observed in consecutive sections) and lonely cells (LC; observed once), illustrating redundancy and overlap.
The right stacked bar shows coverage of \emph{true unique cell IDs} relative to the dense reference, decomposed into captured (green) and missed (red) cells.
As spacing increases, redundancy (SC) decreases and the fraction of missed unique cells increases, reducing overall coverage.
(\textbf{b}) In-plane 3D centroid localization error at $\Delta z=4\,\mu$m relative to the 2~$\mu$m reference.
The histogram shows the distribution of centroid errors.
The dashed black line marks the mean error (2.99~$\mu$m), and the dotted black line marks the standard deviation (3.86~$\mu$m).
The purple dashed line indicates a representative neutrophil diameter for biological context.
Localization errors remain small relative to typical cell size, indicating that sparse reconstruction preserves neighborhood-scale geometry under moderate spacing.
}
\label{fig:eval}
\afterfigure
\end{figure}
}

\newcommand{\figStructures}{
\begin{figure}[t]
\centering
\includegraphics[width=\linewidth]{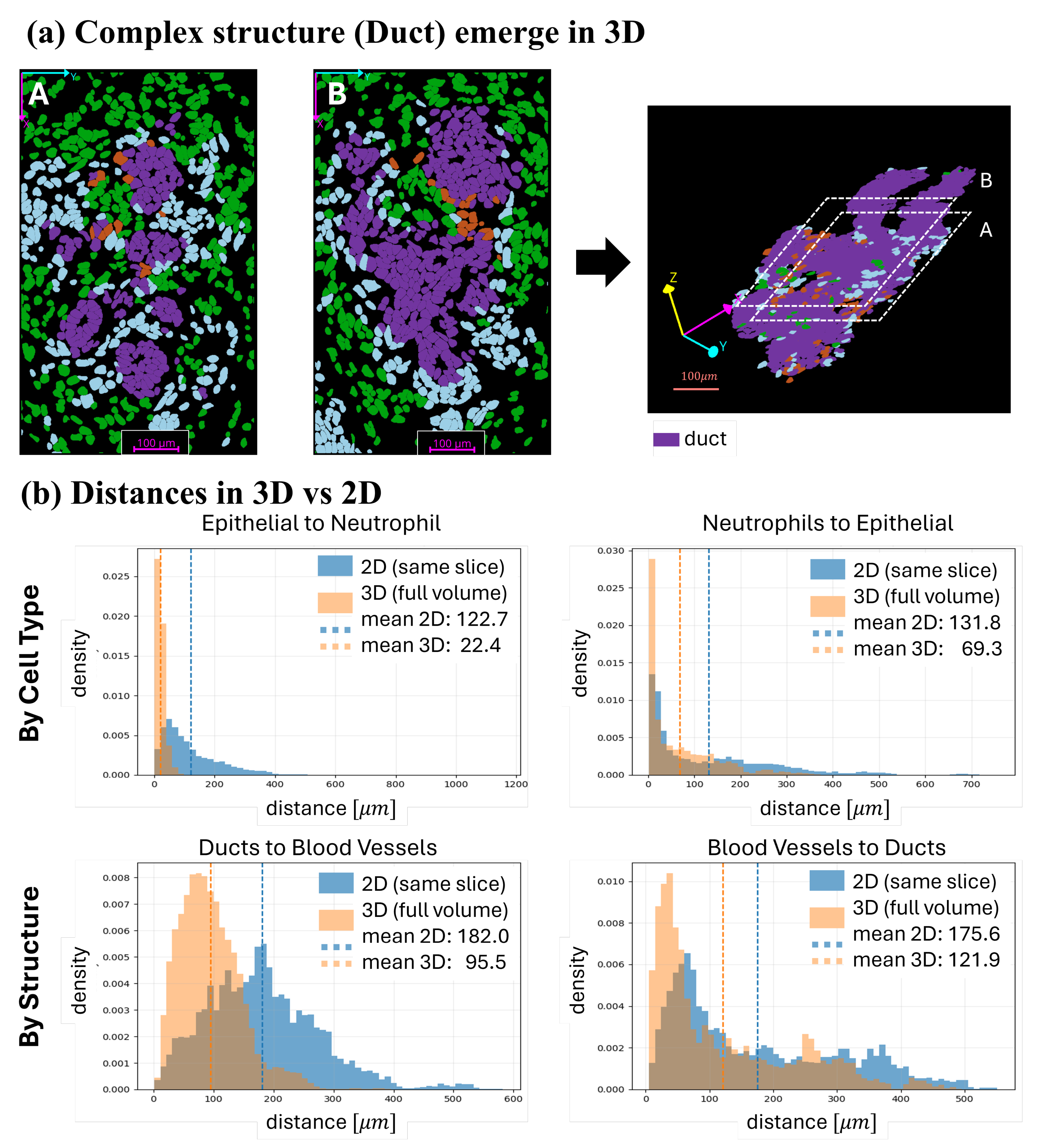}
\vspace*{-0.5cm}
\caption{
\textbf{Structure-centric analysis enabled by sparse 3D reconstruction (PDAC CODEX, $\Delta z=4\,\mu$m).}
(\textbf{a}) Consecutive 2D sections show fragmented cross-sections of a ductal structure (left), which become a coherent connected object after 3D reconstruction (right).
This restoration of continuity enables structure-centric coordinate systems and object-level queries that are not observable in single sections.
(\textbf{b}) Effect of planar measurement on proximity estimates.
For representative cell-type and structure-to-structure queries, distances computed within a single 2D section are systematically larger than distances computed in the reconstructed 3D volume.
Mean distances (dashed lines) shift substantially (e.g., duct--vessel and epithelial--neutrophil pairs), indicating that planar measurements can bias proximity- and interaction-based metrics.
\afterfigure}
\label{fig:structures}
\end{figure}
}

\newcommand{\figGradients}{
\begin{figure}[t]
\centering
\includegraphics[width=\linewidth]{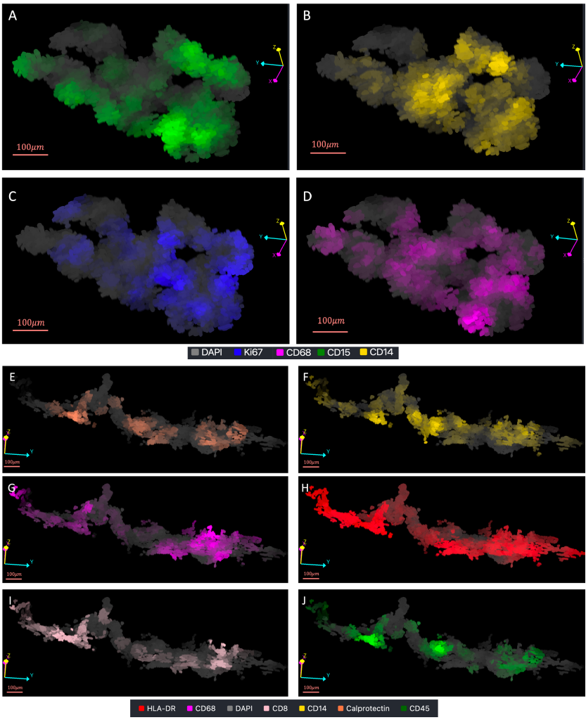}
\caption{\textbf{Along-structure gradients enabled by sparse 3D.}
Using structure-centric coordinates derived from reconstructed 3D objects, we compute marker profiles along ducts/vessels. Sparse 3D reconstruction reduces confounds from arbitrary 2D intersection geometry, enabling depth-aware aggregation of along-structure trends.
\afterfigure}
\label{fig:gradients}
\end{figure}
}

\newcommand{\tabRuleOfThumb}{
\begin{table*}[t]
\centering
\small
\setlength{\tabcolsep}{4pt}
\renewcommand{\arraystretch}{1.15}
\begin{tabular}{p{.17\linewidth} p{.21\linewidth} p{.26\linewidth} p{.3\linewidth}}
\hline
\textbf{Biological goal} & \textbf{Target statistic} & \textbf{Risk under 2D} & \textbf{Preferred acquisition (fixed budget)}   \\
\hline
Abundance / composition & global frequencies, proportions & low (averages out) &
2D sections (maximize coverage)  \\
\hline
Rare population detection & presence / prevalence & medium (cluster-dependent) &
mixed: 2D for discovery; add sparse serial if localized  \\
\hline
Cell--cell interactions & neighborhood enrichment, contact proxies & high (depth collapse confounds neighborhoods) &
sparse serial sections + reconstruction (preserve continuity)  \\
\hline
Spatial clustering / niches & local micro-environments, patches & high (section-to-section variance) &
sparse serial sections + reconstruction  \\
\hline
Structure-centric analysis & ducts/vessels as contiguous objects & very high (fragmentation) &
sparse serial sections + reconstruction  \\
\hline
Along-structure gradients & marker profiles along objects & very high (orientation + depth collapse) &
sparse serial sections + reconstruction  \\
\hline
\end{tabular}
\vspace{3pt}
\caption{\textbf{Decision support under a fixed imaging budget.}
Independent 2D sections are efficient for global composition, while analyses that rely on local neighborhoods, interactions, or extended structures degrade under depth collapse and fragmentation, favoring sparse serial acquisition and reconstruction.
\afterfigure}
\label{tab:decision}
\end{table*}
}
\definecolor{cvprblue}{rgb}{0.21,0.49,0.74}
\usepackage[pagebackref,breaklinks,colorlinks,allcolors=cvprblue]{hyperref}

\def\paperID{6} 
\def\confName{CVPR}
\def\confYear{2026}

\title{Sampling-Aware 3D Spatial Analysis in Multiplexed Imaging}

\author{Ido Harlev \quad Tamar Oukhanov \quad Raz Ben-Uri \quad Leeat Keren \quad Shai Bagon\\
Weizmann Institute of Science, Israel\\
{\tt\small \{ido.harlev, tamar.oukhanov, raz.ben-uri, leeat.keren, shai.bagon\}@weizmann.ac.il}
}

\begin{document}
\figTeaser  
\maketitle
\begin{abstract}
Highly multiplexed microscopy enables rich spatial characterization of tissues at single-cell resolution, yet most analyses rely on two-dimensional sections despite inherently three-dimensional tissue organization. Acquiring dense volumetric data in spatial proteomics remains costly and technically challenging, leaving practitioners to choose between 2D sections or 3D serial sections under limited imaging budgets. In this work, we study how sampling geometry impacts the stability of commonly used spatial statistics, and we introduce a geometry-aware reconstruction module that enables sparse yet consistent 3D analysis from serial sections.
Using controlled simulations with known ground truth, we show that planar sampling reliably recovers global cell-type abundance but exhibits high variance for local statistics such as cell clustering and cell--cell interactions, particularly for rare or spatially localized populations. We observe consistent behavior in real multiplexed datasets, where interaction metrics and neighborhood relationships fluctuate substantially across individual sections. To support sparse 3D analysis in practice, we present a reconstruction approach that links cell projections across adjacent sections using phenotype and proximity constraints and recovers single-cell 3D centroids using cell-type-specific shape priors. We further analyze the trade-off between section spacing, coverage, and redundancy, identifying acquisition regimes that maximize reconstruction utility under fixed imaging budgets.
We validate the reconstruction module on a public imaging mass cytometry dataset with dense axial sampling and demonstrate its downstream utility on an in-house CODEX dataset by enabling structure-level 3D analyses that are unreliable in 2D. Together, our results provide diagnostic tools and practical guidance for deciding when 2D sampling suffices and when sparse 3D reconstruction is warranted in spatial proteomics.
\end{abstract}
 
\section{Introduction}
Highly multiplexed imaging technologies such as CODEX~\cite{goltsev2018codex} and imaging mass cytometry (IMC) enable spatially resolved profiling of dozens of molecular markers at single-cell resolution. These approaches form a core substrate for spatial proteomics, where downstream questions increasingly involve not only \emph{which} cell types are present but also \emph{how} they organize locally and across the tissue~\cite{elhanani2023spatial}. Despite this inherently three-dimensional (3D) organization, most spatial analyses are still performed on individual two-dimensional (2D) sections.

Dense volumetric acquisition in spatial proteomics remains expensive and technically demanding. Recent efforts demonstrate the biological value of densely sampled 3D reconstructions~\cite{kuett2022imc,lin2023multiplexed,kiemen2022coda}, but these settings are often impractical for routine studies. In practice, a common constraint is a \emph{fixed imaging budget}: the experimenter must choose between (i) one 2D section to maximize coverage, or (ii) serial sections to preserve partial depth continuity. This choice is frequently made heuristically, yet it directly governs the reliability of downstream statistics. Figure~\ref{fig:teaser} summarizes our setting, findings, and scope.

This paper advances a vision-centric view of the problem: sampling geometry changes which spatial relationships are observable, and therefore changes estimator stability for common spatial statistics. 
We formalize spatial proteomics under limited acquisition as a structured subsampling problem over a Markov random field, and analyze estimator stability and geometric distortion induced by projection.
We show that global composition (e.g., cell-type abundance) can be stable under planar sampling, whereas local statistics (e.g., neighborhood enrichment and interaction proxies) can exhibit high variance due to depth collapse~-- the loss of along-$z$ neighborhood context when analysing isolated 2D sections~-- and section-to-section variability. We then introduce a lightweight reconstruction \emph{module} that operates downstream of standard preprocessing (whole-slice alignment, segmentation, and cell typing) to enable sparse but consistent 3D point representations for depth-aware analysis.

\noindent{}Our contributions are:
\begin{enumerate}[leftmargin=*, noitemsep, topsep=0pt, parsep=0pt, label=(\roman*)]
\item A controlled study using simulated 3D tissues that isolates how sampling geometry affects recovery of global versus local spatial properties.
\item An empirical stability analysis on a densely sampled IMC dataset showing analogous sampling-induced variability in real data.
\item A geometry-aware reconstruction module with an explicit coverage--redundancy trade-off governed by section spacing.
\item Demonstrations of structure-level, depth-aware analyses enabled by 3D point clouds.
\end{enumerate}

\section{Related Work}
\textbf{3D spatial proteomics and volumetric atlases.}
Volumetric reconstructions from densely sampled serial sections have shown that 3D organization can reveal neighborhoods, gradients, and extended structures that are difficult to interpret in isolated sections~\cite{kuett2022imc,lin2023multiplexed}. Scalable pipelines for large-tissue reconstruction further emphasize the promise of dense 3D integration~\cite{kiemen2022coda}. These works establish what becomes possible when dense 3D data are available, but they largely assume costly acquisition regimes. Our focus is complementary: we analyze \emph{sampling-limited} settings and provide tools for deciding when serial sampling is warranted.

\noindent\textbf{Spatial statistics and cell--cell interactions in multiplexed imaging.}
A broad set of methods quantify spatial organization in multiplexed imaging via neighborhood enrichment, clustering, and interaction metrics~\cite{schapiro2017histocat,dries2021giotto,keren2018structured,forjaz2025three}. Sensitivity of interaction analysis to sampling and heterogeneity has been noted in practice~\cite{arnol2019shifted,jackson2020singlecell}. We build on this literature by explicitly quantifying how sampling geometry (independent 2D vs sparse serial 3D) affects stability of these commonly used statistics.

\noindent\textbf{Reconstruction and correspondence across sections.}
Serial-section reconstruction often relies on dense sampling, pixel-level registration, or joint optimization across multiple stages~\cite{kiemen2022coda}. Object correspondence and assignment problems are also central in computer vision and cell tracking~\cite{kuhn1955hungarian,maska2023celltracking}. We adopt a constrained assignment view and propose a lightweight reconstruction module that targets sparse serial acquisition and operates downstream of segmentation and cell typing.
\vspace*{-2mm}
\section{Problem Setting and Sampling Geometry}
We represent a tissue sample as a set of segmented cells with spatial coordinates and discrete cell-type labels. Typical spatial proteomics analyses compute (i)~global statistics such as abundance and composition, and (ii)~local statistics such as neighborhood enrichment and interaction proxies based on spatial proximity graphs~\cite{schapiro2017histocat,dries2021giotto}.

Under a fixed imaging budget, two acquisition geometries are common. \emph{2D sampling} acquires a single section, maximizing coverage but discarding depth continuity. \emph{3D serial sampling} acquires sections at regular axial spacing $\Delta z$, preserving partial depth continuity but covering reduced area. These choices induce different estimator behavior: global composition averages over observations and can be robust, while local statistics depend on intact neighborhood structure and are sensitive to depth collapse~-- the missing along-$z$ context in isolated 2D sections.

In the next sections we quantify these effects using controlled simulations (Sec.~\ref{sec:sim}) and real Imaging Mass Cytometry (IMC) data (Sec.~\ref{sec:imc}), then introduce a reconstruction module designed specifically for serial sampling (Sec.~\ref{sec:method}).

\section{Controlled Analysis with Simulated Data}
\label{sec:sim}
\figSim
To isolate the effect of sampling geometry from biological variability, we construct simulated 3D tissues with \emph{known} global composition and local spatial organization, and evaluate how well these properties can be recovered from matched-budget observations under independent 2D sampling versus serial sampling (Fig.~\ref{fig:sim}). The simulation is based on a pairwise Markov random field (MRF), a standard model for spatial processes and graphical models~\cite{koller2009pgm,wainwright2008graphical,ripley1988spatial}.

\subsection{MRF tissue model}
Let $G=(V,E)$ be a 3D lattice graph, where each site $i\in V$ corresponds to a voxel/``cell" and edges $(i,j)\in E$ connect spatial neighbors (e.g., 26-neighborhood). Each voxel has a discrete label $x_i \in \{1,\dots,K\}$ representing a cell type. We define a Gibbs distribution
\begin{equation}
p(\mathbf{x}\mid \boldsymbol{\alpha}, \mathbf{B})
\;=\;
\frac{1}{Z(\boldsymbol{\alpha},\mathbf{B})}
\exp\!\Big(
\sum_{i\in V} \alpha_{x_i}
+
\sum_{(i,j)\in E} B_{x_i,x_j}
\Big),
\label{eq:mrf}
\end{equation}
where $\boldsymbol{\alpha}\in\mathbb{R}^K$ are unary (``field'') parameters controlling global prevalence, $\mathbf{B}\in\mathbb{R}^{K\times K}$ encodes pairwise affinities between types, and $Z$ is the partition function. Fig.~\ref{fig:sim}\textbf{(a-b)} exemplifies these two components: $\boldsymbol{\alpha}$ sets the baseline prevalence of each label, while $\mathbf{B}$ biases which cell-types prefer to be adjacent or apart. The diagonal of $\mathbf{B}$ determines the tendency of each cell-type to self-cluster.

Larger $B_{ab}$ increases the probability that types $a$ and $b$ are neighbors, inducing clustered domains and structured neighborhoods; diagonal terms act as within-type cohesion. This effect is visible in Fig.~\ref{fig:sim}\textbf{(a)}: with comparable global prevalence, changing $\mathbf{B}$ can produce either near-i.i.d.\ mixtures or strongly clustered spatial organizations. On the other hand, fixing $\mathbf{B}$ and changing the global prevalence (Fig.~\ref{fig:sim}\textbf{(b)}) leads to scaling effects.

In this parameterization, the \emph{global} property of interest is prevalence (captured by $\boldsymbol{\alpha}$ and the implied marginals), whereas the \emph{local} property of interest is interaction structure (captured by $\mathbf{B}$), which governs neighborhood composition and proximity-graph interaction proxies.

\subsection{Generation of synthetic tissues}
For each simulation regime, we choose $(\boldsymbol{\alpha},\mathbf{B})$ to yield a target composition and interaction structure, including cases with rare and spatially localized populations. We then sample tissue configurations $\mathbf{x}$ from~\eqref{eq:mrf} using standard MRF sampling procedures. 

\subsection{Sampling geometries under a matched budget}
From each simulated 3D volume, we extract observations under a \emph{matched voxel budget} using two geometries:
\begin{itemize}
\item \textbf{Independent 2D sampling:} $M$ independent planar sections are selected at diverse depths; each yields a 2D lattice subgraph with labels to ensure heterogeneity.
\item \textbf{Serial sampling:} a stack of serial sections is selected at spacing $\Delta z$, yielding partial depth continuity (a thin 3D subgraph).
\end{itemize}
Both strategies observe the same number of voxels, so differences in recovery are attributable to geometry rather than sample count. In Fig.~\ref{fig:sim}\textbf{(c)}, we compare these strategies against the full simulated volume (``FullSim'') as a best-case reference.

\subsection{Parameter recovery via maximum pseudo-likelihood}
Exact maximum likelihood estimation for MRFs is intractable at this scale due to the partition function $Z(\boldsymbol{\alpha},\mathbf{B})$. We therefore estimate parameters using maximum pseudo-likelihood (MPLE)~\cite{besag1975statistical}, which maximizes the product of node-wise conditional probabilities. For a node $i$ with neighbor set $\mathcal{N}(i)$, the conditional distribution induced by~\eqref{eq:mrf} is
\begin{equation}
p(x_i=a \mid \mathbf{x}_{\mathcal{N}(i)}, \boldsymbol{\alpha},\mathbf{B})
\;=\;
\frac{
\exp\!\big(\alpha_a + \sum_{j\in\mathcal{N}(i)} B_{a,x_j}\big)
}{
\sum_{b=1}^K \exp\!\big(\alpha_b + \sum_{j\in\mathcal{N}(i)} B_{b,x_j}\big)
}.
\label{eq:conditional}
\end{equation}
The MPLE objective is then
\begin{equation}
\hat{\boldsymbol{\alpha}},\hat{\mathbf{B}}=\arg\max_{\boldsymbol{\alpha},\mathbf{B}}
\;\sum_{i\in \Omega}
\log p(x_i \mid \mathbf{x}_{\mathcal{N}(i)}, \boldsymbol{\alpha},\mathbf{B})
\;-\;\lambda \lVert \mathbf{B}\rVert_F^2,
\label{eq:mple}
\end{equation}
where $\Omega$ indexes observed nodes under the sampling geometry, and $\lambda$ weighs the regularization term. Intuitively, $\boldsymbol{\alpha}$ is driven by global counts, whereas $\mathbf{B}$ is driven by reliable observation of neighborhoods; Fig.~\ref{fig:sim}\textbf{(c)} anticipates this separation by showing that prevalence-related errors remain small while interaction-related errors inflate under planar sampling.

\subsection{Findings: global versus local recovery}
Across regimes, we evaluate recovery of (i)~global prevalence via $\hat{\boldsymbol{\alpha}}$ and (ii)~interaction structure via $\hat{\mathbf{B}}$, using MAE/RMSE relative to the ground-truth parameters used to generate each sample. The key pattern in Fig.~\ref{fig:sim}\textbf{(c)} is consistent: prevalence is recovered reliably under both sampling strategies, while interaction recovery degrades sharply under independent 2D sampling and exhibits higher variance across repeated draws, especially for rare or spatially localized populations. In contrast, serial sampling preserves enough neighborhood continuity to stabilize estimates of $\mathbf{B}$ under the same observation budget.


These controlled results motivate our empirical stability analysis on real IMC volumes (Sec.~\ref{sec:imc}) and the development of a reconstruction module to support depth-aware spatial analysis under serial sampling (Secs.~\ref{sec:method}--\ref{sec:eval}).

\section{Empirical Stability in Real Data}
\label{sec:imc}
\figReal
We next test whether the simulation-derived behavior manifests in real multiplexed microscopy under dense axial sampling. We analyze a public 3D IMC dataset with 2~$\mu$m spacing from Kuett~\etal~\cite{kuett2022imc}, which we treat as an approximate volumetric reference. Figure~\ref{fig:imc} summarizes three section-based quantities~-- abundance, detectability, and neighborhood enrichment~-- computed across depth, mirroring common analysis workflows that operate on individual sections~\cite{schapiro2017histocat,dries2021giotto}.

\paragraph{Global abundance (what exists in the volume).}
We begin by quantifying global cell-type abundance in the full reconstructed volume (Fig.~\ref{fig:imc}\textbf{(a)}). This panel establishes the baseline prevalence of each population and motivates a natural question: under planar analysis, which of these populations can be \emph{reliably observed} from a limited number of sections?

\paragraph{Detectability under planar sampling (what you actually ``see'' in 2D).}
To translate global abundance into expected observability under sectioning, we compute a detectability statistic: for each cell type, we repeatedly sample $M{=}20$ sections and measure the fraction of trails in which at least $k{=}100$ cells of that type are observed (Fig.~\ref{fig:imc}\textbf{(b)}). This quantity is intentionally section-centric: it captures not only rarity (low global abundance) but also spatial organization. For example, B cells show low detectability, consistent with low abundance and high spatial-clustering; however, endothelial cells can be detected reliably despite comparatively low abundance, reflecting spatial dispersion across the tissue. Conversely, some fibroblast subtypes remain difficult to detect consistently despite moderate-to-high abundance, indicating strong spatial clustering that causes many sections to miss the relevant regions entirely. This demonstrates that ``how often a type is seen'' depends jointly on prevalence and 3D clustering, not prevalence alone.

\paragraph{Local spatial statistics (what relationships you infer).}
We then evaluate a standard interaction proxy based on neighborhood enrichment~\cite{keren2018structured,arnol2019shifted}: for a fixed target type (here Fibroblasts\_apCAF), we compute across sections the enrichment z-score of observing each partner type within a fixed radius, relative to a label-permutation null (Fig.~\ref{fig:imc}\textbf{(c)}). The resulting distributions show that enrichment estimates fluctuate substantially with section choice: for some partners, the same tissue volume yields sections with strong apparent enrichment (beyond $|z|{=}2$) and others with weak or negligible association. In practical terms, section choice can therefore change whether a given interaction is interpreted as ``present'' or ``absent,'' even before considering biological variability between patients.

Together, Fig.~\ref{fig:imc} shows that planar sampling induces high variance in local spatial statistics within a single tissue volume, while global composition is comparatively stable. This motivates acquisition-aware reasoning (Tab.~\ref{tab:decision}) and, when local spatial questions are central, sparse serial sampling with reconstruction 
to stabilize depth-sensitive measurements.

\section{Sparse 3D Reconstruction Module}
\label{sec:method}
\figMatching

To enable depth-aware spatial analysis under sparse serial acquisition, we reconstruct a sparse 3D point cloud of cells by linking projections across adjacent sections (Fig.~\ref{fig:method}). The objective is not dense volumetric imaging, but consistent 3D localization sufficient to preserve neighborhood structure across depth.

\noindent\textbf{Scope and inputs.}
We assume that whole-slice alignment, segmentation, and cell-type classification have already been performed using established tools (e.g.,~\cite{goldsborough2024instanseg,bussi2025celltune}). Our module operates downstream of these steps and focuses on inter-slice correspondence and 3D centroid estimation. This scoped formulation distinguishes our approach from dense end-to-end reconstruction pipelines such as CODA~\cite{kiemen2022coda}.

\noindent\textbf{Inter-slice correspondence as constrained assignment.}
Figure~\ref{fig:method}\textbf{(a)} illustrates the matching challenge in real PDAC CODEX data acquired at $\Delta z\!=\!4\,\mu$m. 
Because cells are intersected by successive cutting planes, a single biological cell may appear as multiple cross-sections 
in adjacent slices. However, multiple nearby candidates of the same phenotype may also exist, making correspondence ambiguous.

To formalize matching, we compute the in-plane Euclidean distance matrix $D_{ij}$ between centroids in sections $z$ and $z\!+\!\Delta z$ (Fig.~\ref{fig:method}\textbf{(b)(ii)}). 
We impose two constraints:
\begin{itemize}
\item \textbf{Phenotype constraint:} pairs with non-matching cell-type labels are assigned $D_{ij}=\infty$.
\item \textbf{Proximity constraint:} pairs exceeding a cell-type-specific tolerance (derived from empirical size statistics) are rejected.
\end{itemize}



The resulting sparse cost matrix is solved using the Hungarian algorithm~\cite{kuhn1955hungarian} to obtain a one-to-one assignment that minimizes total in-plane displacement while allowing cells to remain unmatched. 
Matching across sections incurs a displacement cost, whereas leaving a cell unmatched carries an implicit penalty; the optimizer therefore balances forming cross-section correspondences against forcing implausible matches.
Selected correspondences are highlighted in Fig.~\ref{fig:method}\textbf{(b)(ii)}.

This procedure partitions the segmented cells into two regimes (Fig.~\ref{fig:method}\textbf{(b)(iii)}): 
\emph{shared cells} (SC), matched across adjacent sections and interpreted as the same biological cell, and 
\emph{lone cells} (LC), observed in only one section with no valid correspondence. 
The relative frequency of these regimes depends directly on section spacing, linking matching behavior to acquisition design.

\noindent\textbf{Geometry-aware centroid estimation.}
Matching alone provides adjacency correspondences but not depth localization.
We therefore introduce a weak geometric prior based on the observation that cells can be approximated as ellipsoids whose cross-sectional area varies with the cutting plane offset.

As shown in Fig.~\ref{fig:method}\textbf{(c)}, empirical distributions of 2D cross-sectional areas differ substantially across phenotypes (e.g., macrophage-Calp, SMV, CD8 T-cells, tumor). 
These distributions are used to estimate cell-type-specific size statistics, which serve two purposes:
(i) defining proximity tolerances during matching, and
(ii) regularizing depth inference during centroid estimation.

For SC, multiple cross-sections across adjacent planes constrain the centroid position under the ellipsoidal model.
For LC, in-plane coordinates are retained while depth is bounded between neighboring section planes.
The resulting sparse 3D point cloud preserves cell identity and approximate spatial relationships across depth, enabling downstream depth-aware analyses (Secs.~\ref{sec:eval}--\ref{sec:downstream}).

\section{Reconstruction and Sampling Trade-offs}
\label{sec:eval}
\figMatchingRes

We evaluate the reconstruction module using the densely sampled IMC dataset of Kuett~\etal~\cite{kuett2022imc} by treating the 2~$\mu$m stack as a pseudo-volumetric reference and subsampling sections at increasing axial spacing $\Delta z$. This allows us to quantify how coverage and localization accuracy degrade as sampling becomes sparser, compared to the original annotations of~\cite{kuett2022imc} done on the densely sampled data.

\noindent\textbf{Coverage as a function of spacing.}
Figure~\ref{fig:eval}\textbf{(a)} decomposes the effect of axial spacing $\Delta z$ relative to the dense 2~$\mu$m reference.
For each spacing, the left stacked bar shows the composition of cells observed in the sampled stack, separated into shared cells (SC) and lonely cells (LC), thereby quantifying redundancy across adjacent sections.
The right stacked bar shows coverage of true unique cell IDs relative to the dense reference, decomposed into captured and missed cells.

At 2~$\mu$m, redundancy dominates: most cells are observed in consecutive sections (high SC fraction), and coverage of unique IDs is complete by definition.
At 4~$\mu$m, global unique-cell coverage remains high (92.6\%), while the SC fraction is reduced, indicating less overlap but minimal loss of unique cells.
As spacing increases to 6--10~$\mu$m, the SC fraction declines further and the proportion of missed unique cells grows, reflecting an increasing number of non-captured cells between planes.

This decomposition makes the redundancy--coverage trade-off explicit: dense spacing yields high overlap (many SC cells), whereas larger $\Delta z$ reduces overlap and increases the probability of missing cells entirely.
Moderate spacing (e.g., 4~$\mu$m) therefore retains most unique cells while reducing acquisition redundancy.
The optimal spacing depends on cell size distributions and the spatial statistics of interest.




\noindent\textbf{Localization accuracy.} 
Beyond coverage, we evaluate the 3D localization error of reconstructed centroids relative to the 2~$\mu$m reference.
Figure~\ref{fig:eval}\textbf{(b)} reports the distribution of centroid errors at $\Delta z=4\,\mu$m.
The mean localization error is 2.99~$\mu$m (std 3.86~$\mu$m), substantially smaller than representative cell diameters (e.g., $\sim$8~$\mu$m for Neutrophils~\cite{Rosenbluth2006neut_size}).

This comparison to biological scale is critical: although spacing increases introduce uncertainty in depth estimation, the resulting errors remain a fraction of typical cell size.
Consequently, neighborhood-scale spatial relationships are largely preserved under sparse reconstruction in this regime.

\noindent\textbf{Implications for acquisition design.}
Taken together, Fig.~\ref{fig:eval} shows that section spacing acts as a controllable design parameter.
Very dense sampling (2~$\mu$m) yields maximal redundancy and negligible coverage loss but incurs higher imaging cost.
Moderate spacing (e.g., 4~$\mu$m) retains most unique cells and maintains localization error well below cell diameter, providing a practical operating point under fixed imaging budgets.
Larger spacings further reduce acquisition effort but increasingly compromise coverage and correspondence reliability.

These empirical results support the view that sparse serial sampling, combined with geometry-aware reconstruction, can recover meaningful 3D structure without requiring extremely dense axial acquisition.

\tabRuleOfThumb
\section{Structure-level analyses enabled by 3D}
\label{sec:downstream}
\figStructures

We next illustrate how our 3D reconstruction changes the geometry of downstream spatial analyses in a PDAC CODEX dataset acquired at $\Delta z=4\,\mu$m. The focus here is not biological discovery, but how restoring depth continuity alters measurable quantities.

\paragraph{From fragmented slices to coherent 3D objects}
Figure~\ref{fig:structures}\textbf{(a)} shows two consecutive sections in which a duct appears as spatially disconnected regions.
When reconstructed in 3D, these fragments form a single connected structure.
This transition from disconnected cross-sections to a coherent object enables structure-centric representations, such as defining a surface, a centerline, or a local coordinate system along the duct.
Such representations are ill-posed when analyzing individual sections independently.

\paragraph{Depth collapse distorts distance measurements}
Fig.~\ref{fig:structures}\textbf{(b)} quantifies the geometric consequences of depth collapse.
For representative cell-type and structure-level queries, we compare distances measured within a single section (2D) to distances computed in the reconstructed volume (3D).
Across all shown cases, 2D distances are systematically larger than 3D distances.
For example, duct-to-vessel and epithelial-to-neutrophil separations exhibit substantial shifts in mean distance when depth information is incorporated.
This discrepancy arises because planar measurements discard out-of-plane proximity: cells that are close in 3D may appear far apart in a single section.

These results demonstrate that sparse 3D reconstruction does not merely ``visualize'' tissue differently, but changes the numerical values of proximity-based metrics.
Consequently, interaction scores and distance-to-structure analyses derived from single sections can be biased by the missing along-$z$ sections.




\figGradients
\paragraph{Along-structure gradients under sparse sampling}
Finally, we compute marker profiles along reconstructed structures (ducts/vessels) using the 3D object coordinate system (Fig.~\ref{fig:gradients}). Such analyses are sensitive to orientation and depth; in single sections, profiles are confounded by arbitrary intersection geometry. Sparse reconstruction enables depth-aware aggregation along extended structures, providing a principled mechanism to study gradients under practical sampling.

\section{Practical Guidelines, Limitations, and Outlook}
Our results support a simple principle: under fixed budgets, acquisition geometry should be matched to the statistic of interest. 2D sampling maximizes coverage and stabilizes global composition estimates, while analyses that depend on local neighborhoods, interaction proxies, or extended structures benefit from depth continuity and favor sparse serial sampling with reconstruction. Tab.~\ref{tab:decision} summarized these considerations into a set of practitioner's rules-of-thumb.

Our reconstruction module assumes accurate whole-slice alignment and reliable cell typing, and correspondence degrades under severe crowding or phenotype ambiguity. The geometric prior is intentionally weak; it supports depth estimation for shared cells but does not capture complex morphologies. These limitations suggest straightforward extensions: incorporate learned correspondence models, integrate 3D-aware typing, and jointly quantify how reconstruction uncertainty propagates into downstream spatial statistics.
Our study also provides empirical guidelines for informed selection of axial spacing when sampling 3D serial sections.

More broadly, we view sampling-aware analysis (Secs.~\ref{sec:sim}--\ref{sec:imc}) and sparse reconstruction (Secs.~\ref{sec:method}--\ref{sec:eval}) as complementary tools that bridge current acquisition constraints and the increasing demand for depth-aware spatial understanding in multiplexed microscopy.

\paragraph{Acknowledgments:}
This project received funding from
MBZUAI-WIS Joint Program for AI Research,
and the Knell Family Institute for Artificial Intelligence.
L.K. holds the Fred and Andrea Fallek President's Development Chair. She is supported by the Enoch foundation research fund, Fundación Alberto Palatchi, the Abisch-Frenkel foundation, the Rising Tide foundation, the Sharon Levine Foundation, the Rosetrees foundation and grants funded by the Dwek center for cancer immunotherapy, European Research Council (948811), the Israel Science Foundation (2481/20, 3830/21) within the Israel Precision Medicine Partnership program and the Melanoma Research Aliance Team Science Award (1200724).

\newpage
{
    \small
    \bibliographystyle{ieeenat_fullname}
    \bibliography{main}
}


\end{document}